\def\year{2020}\relax
\documentclass[letterpaper]{article} 
\usepackage{aaai20}  
\usepackage{times}  
\usepackage{helvet} 
\usepackage{courier}  
\usepackage[hyphens]{url}  
\usepackage{graphicx} 

\usepackage[caption=false]{subfig}
\usepackage{float}
\renewcommand{\vec}[1]{\mathbf{#1}}
\usepackage{amsmath}
\usepackage[dvipsnames]{xcolor}
\usepackage{graphicx}
\usepackage{wrapfig}
\usepackage{float,lipsum,multicol} 
\usepackage{lineno}
\usepackage{graphicx}
\usepackage{booktabs}
\usepackage{amssymb,amsmath,nccmath}
\usepackage{cclicenses}
\usepackage{makecell}
\usepackage{lscape,array}
\newcolumntype{C}[1]{>{\centering\arraybackslash}p{#1}} 
\usepackage[thin, , thinc]{esdiff}
\usepackage{caption}
\usepackage{framed}  
\usepackage[font=small,skip=0pt]{caption}
\usepackage{algorithmicx,algorithm}
\usepackage[noend]{algpseudocode}

\usepackage{multirow}
\usepackage{makecell}
\usepackage{graphicx}  
\title{AutoCompress: An Automatic DNN Structured Pruning Framework for Ultra-High Compression Rates }

\author{\fontsize{11}{10}\selectfont Ning Liu{$^\dagger$}\textsuperscript{1}, Xiaolong Ma\textsuperscript{1}, Zhiyuan Xu\textsuperscript{3}, Yanzhi Wang\textsuperscript{1}, Jian Tang\textsuperscript{2,3}, Jieping Ye\textsuperscript{2,4}\\
\textsuperscript{1}Northeastern University, 
\textsuperscript{2}DiDi AI Labs, 
\textsuperscript{3}Syracuse University, 
\textsuperscript{4}University of Michigan \\
{E-mail:}\{liu.ning, ma.xiaol\}@husky.neu.edu,
zxu105@syr.edu,\\
yanz.wang@northeastern.edu,
jtang02@syr.edu,
yejieping@didiglobal.com
}

\begin{document}
\maketitle
\vspace{-0.1in}
\begin{abstract}
Structured weight pruning is a representative model compression technique of DNNs to reduce the storage and computation requirements and accelerate inference. An automatic hyperparameter determination process is necessary due to the large number of flexible hyperparameters.
This work proposes AutoCompress, an automatic structured pruning framework with the following key performance improvements: (i) effectively incorporate the combination of structured pruning schemes in the automatic process; (ii) adopt the state-of-art ADMM-based structured weight pruning as the core algorithm, and propose an innovative additional purification step for further weight reduction without accuracy loss; and (iii) develop effective heuristic search method enhanced by experience-based guided search, replacing the prior deep reinforcement learning technique which has underlying incompatibility with the target pruning problem.
Extensive experiments on CIFAR-10 and ImageNet datasets demonstrate that AutoCompress is the key to achieve ultra-high pruning rates on the number of weights and FLOPs that cannot be achieved before. As an example, AutoCompress outperforms the prior work on automatic model compression by up to 33$\times$ in pruning rate (120$\times$ reduction in the actual parameter count) under the same accuracy. Significant inference speedup has been observed from the AutoCompress framework on actual measurements on smartphone.
We release all models of this work at anonymous link: http://bit.ly/2VZ63dS.
\end{abstract}

\section{Introduction}
The high computational and storage requirements of large-scale DNNs, such as VGG \cite{Simonyan2015VeryDC} or ResNet \cite{he2016deep}, make it prohibitive for broad, real-time applications at the mobile end. Model compression techniques have been proposed that aim at reducing both the storage and computational costs for DNN inference phase \cite{wen2016learning,luo2017entropy,min20182pfpce,guo2016dynamic,han2015learning,he2017channel,he2018amc,zhang2018systematic,zhang2018adam,rastegari2016xnor,leng2018extremely}. One key model compression technique is DNN \emph{weight pruning} \cite{wen2016learning,luo2017entropy,min20182pfpce,guo2016dynamic,han2015learning,he2017channel,he2018amc,zhang2018systematic,zhang2018adam} that reduces the number of weight parameters, with minor (or no) accuracy loss.

There are mainly two categories of weight pruning. The general, \emph{non-structured pruning} \cite{luo2017entropy,guo2016dynamic,han2015learning,zhang2018systematic} can prune arbitrary weight in DNN. Despite the high pruning rate (weight reduction), it suffers from limited acceleration in actual hardware implementation due to the sparse weight matrix storage and associated indices \cite{han2015learning,wen2016learning,he2017channel}. On the other hand, \emph{structured pruning} \cite{wen2016learning,min20182pfpce,he2017channel,zhang2018adam} can directly reduce the size of weight matrix while maintaining the form of a full matrix, without the need of indices. It is thus more compatible with hardware acceleration and has become the recent research focus. There are multiple types/schemes of structured pruning, e.g., \emph{filter pruning}, \emph{channel pruning}, and \emph{column pruning} for CONV layers of DNN as summarized in \cite{wen2016learning,luo2017entropy,he2017channel,zhang2018adam}. Recently, a systematic solution framework \cite{zhang2018systematic,zhang2018adam} has been developed based on the powerful optimization tool ADMM (Alternating Direction Methods of Multipliers) \cite{boyd2011distributed,suzuki2013dual}. It is applicable to different schemes of structured pruning (and non-structured one) and achieves state-of-art results \cite{zhang2018systematic,zhang2018adam} by far.

The structured pruning problem of DNNs is flexible, comprising a large number of hyper-parameters, including the scheme of structured pruning and combination (for each layer), per-layer weight pruning rate, etc. Conventional hand-crafted policy has to explore the large design space for hyperparameter determination for weight or computation (FLOPs) reductions, with minimum accuracy loss. The trial-and-error process is highly time-consuming, and derived hyperparameters are usually sub-optimal. It is thus desirable to employ an automated process of hyperparameter determination for such structured pruning problem, motivated by the concept of AutoML (automated machine learning) \cite{zoph2016neural,baker2016designing,li2016hyperband,real2017large,liu2018progressive}. Recent work AMC \cite{he2018amc} employs the popular \emph{deep reinforcement learning} (DRL) \cite{zoph2016neural,baker2016designing} technique for automatic determination of per-layer pruning rates. However, it has limitations that (i) it employs an early weight pruning technique based on fixed regularization, and (ii) it only considers filter pruning for structured pruning. As we shall see later, the underlying incompatibility between the utilized DRL framework with the problem further limits its ability to achieve high weight pruning rates (the maximum reported pruning rate in \cite{he2018amc} is only 5$\times$ and is non-structured pruning).

This work makes the following innovative contributions in the automatic hyperparameter determination process for DNN structured pruning. First, we analyze such automatic process in details and extract the \emph{generic flow}, with four steps: (i) \emph{action sampling}, (ii) \emph{quick action evaluation}, (iii) \emph{decision making}, and (iv) \emph{actual pruning and result generation}. Next, we identify three sources of performance improvement compared with prior work. We adopt the ADMM-based structured weight pruning algorithm as the core algorithm, and propose an innovative additional purification step for further weight reduction without accuracy loss. Furthermore, we find that the DRL framework has underlying incompatibility with the characteristics of the target pruning problem, and conclude that such issues can be mitigated simultaneously using effective \emph{heuristic search method} enhanced by experience-based guided search. 

Combining all the improvements results in our automatic framework \textbf{AutoCompress}, which outperforms the prior work on automatic model compression by up to 33$\times$ in pruning rate (120$\times$ reduction in the actual parameter count) under the same accuracy. Through extensive experiments on CIFAR-10 and ImageNet datasets, we conclude that AutoCompress is the key to achieve ultra-high pruning rates on the number of weights and FLOPs that cannot be achieved before, while DRL cannot compete with human experts to achieve high pruning rates. Significant inference speedup has been observed from the AutoCompress framework on actual measurements on smartphone, based on our compiler-assisted mobile DNN acceleration framework.
We release all models of this work at
anonymous link: http://bit.ly/2VZ63dS.

\section{Related Work}
\textbf{DNN Weight Pruning and Structured Pruning:}
DNN weight pruning includes two major categories: the general, \emph{non-structured} pruning \cite{luo2017entropy,guo2016dynamic,han2015learning,zhang2018systematic} where arbitrary weight can be pruned, and \emph{structured} pruning \cite{wen2016learning,luo2017entropy,min20182pfpce,he2017channel,zhang2018adam} that maintains certain regularity. Non-structured pruning can result in a higher pruning rate (weight reduction). However, as weight storage is in a sparse matrix format with indices, it often results in performance degradation in highly parallel implementations like GPUs. This limitation can be overcome in structured weight pruning.

\begin{figure}[!t]
    \centering
    \includegraphics[width=0.4\textwidth]{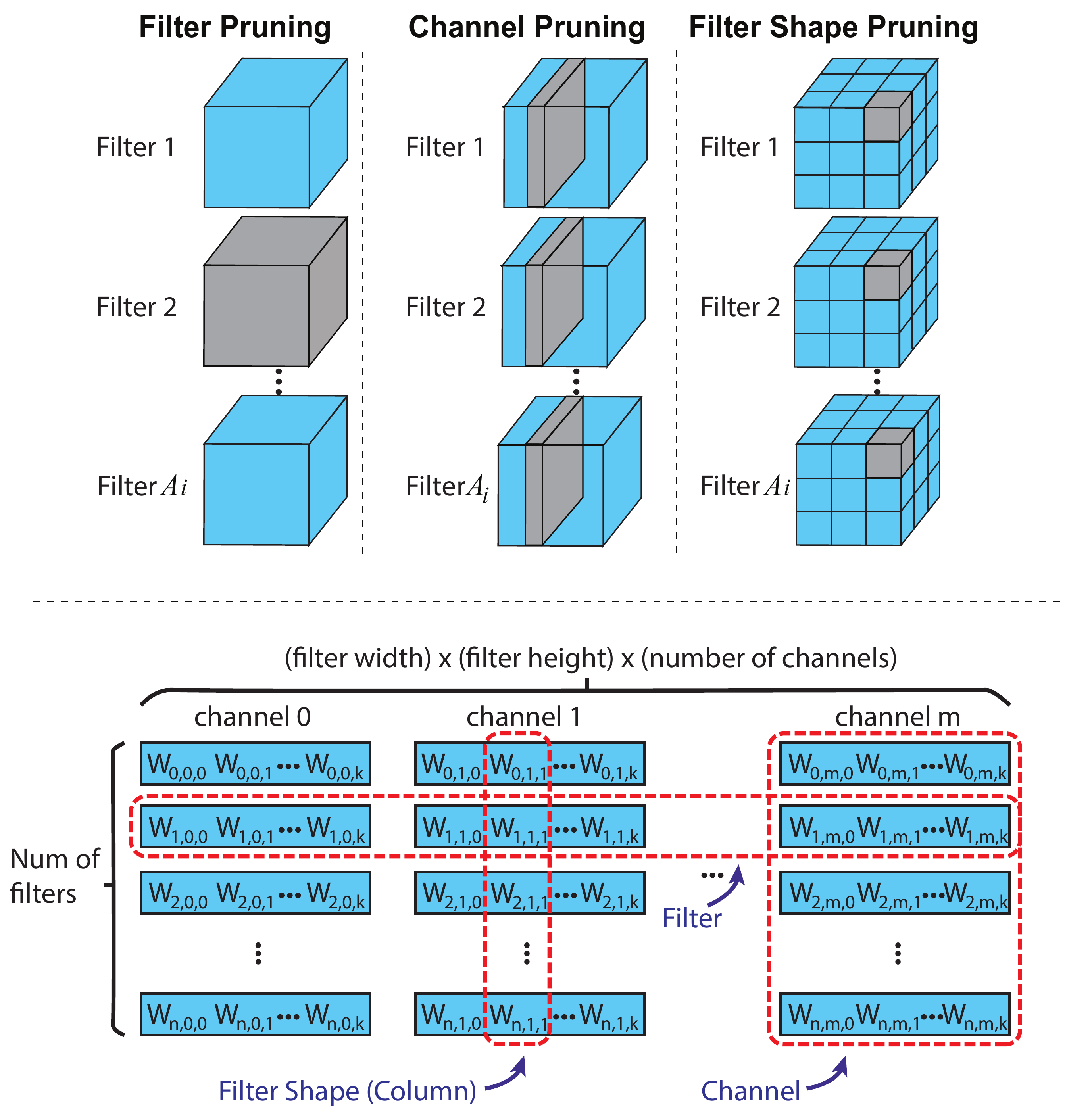}
    \caption{Different structured pruning schemes: A filter-based view and a GEMM view.}
    \label{Fig:gemm}
\vspace{-0.2in}
\end{figure}

Figure~\ref{Fig:gemm} illustrates three structured pruning schemes on the CONV layers of DNN: \emph{filter pruning}, \emph{channel pruning}, and \emph{filter-shape pruning} (a.k.a. \emph{column pruning}), removing whole filter(s), channel(s), and the same location in each filter in each layer. CONV operations in DNNs are commonly transformed to matrix multiplications by converting weight tensors and feature map tensors to matrices \cite{wen2016learning}, named \emph{general matrix multiplication} (GEMM). The key advantage of structured pruning is that a full matrix will be maintained in GEMM with dimensionality reduction, without the need of indices, thereby facilitating hardware implementations.

It is also worth mentioning that filter pruning and channel pruning are correlated \cite{he2017channel}, as pruning a filter in layer $i$ (after batch norm) results in the removal of corresponding channel in layer $i+1$. The relationship in ResNet \cite{he2016deep} and MobileNet \cite{sandler2018mobilenetv2} will be more complicated due to bypass links.

\textbf{ADMM:}
Alternating Direction Method of Multipliers (ADMM) is a powerful mathematical optimization technique, by decomposing an original problem into two subproblems that can be solved separately and efficiently \cite{boyd2011distributed}. Consider the general optimization problem $\min_{\bf{x}} f({\bf x}) + g({\bf x})$. In ADMM, it is decomposed into two subproblems on $\bf x$ and
$\bf z$ ($\bf z$ is an auxiliary variable), to be solved iteratively until convergence. The first subproblem derives $\bf x$ given
$\bf z$: $\min_{\bf x} f({\bf x})+q_{1}({\bf x}|{\bf z})$. The second subproblem derives $\bf z$ given $\bf x$: $\min_{\bf z} g({\bf z})+q_2({\bf z}|{\bf x})$. Both $q_1$ and $q_2$ are quadratic functions. 

As a key property, ADMM can effectively deal with a subset of combinatorial constraints and yield optimal (or at least high quality) solutions. The associated constraints in DNN weight pruning (both non-structured and structured) belong to this subset \cite{hong2016convergence}.
In DNN weight pruning problem, $f({\bf x})$ is loss function of DNN and the first subproblem is DNN training with dynamic regularization, which can be solved using current gradient descent techniques and solution tools \cite{kingma2014adam,TensorFlow-Lite} for DNN training. $g({\bf x})$ corresponds to the combinatorial constraints on the number of weights. As the result of the compatibility with ADMM, the second subproblem has optimal, analytical solution for weight pruning via Euclidean projection. This solution framework applies both to non-structured and different variations of structured pruning schemes.

\textbf{AutoML:} Many recent work have investigated the concept of \emph{automated machine learning} (AutoML), i.e., using machine learning for hyperparameter determination in DNNs.
Neural architecture search (NAS) \cite{zoph2016neural,baker2016designing,liu2018progressive} is an representative application of AutoML. 
NAS has been deployed in Google’s Cloud AutoML framework, which frees customers from the time-consuming DNN architecture design process. 
The most related prior work, AMC \cite{he2018amc}, applies AutoML for DNN weight pruning, leveraging a similar DRL framework as Google AutoML to generate weight pruning rate for each layer of the target DNN.  
In conventional machine learning methods, the overall performance (accuracy) depends greatly on the quality of features. 
To reduce the burdensome manual feature selection process, automated feature engineering learns to generate appropriate feature set in order to improve the performance of corresponding machine learning tools.

\section{The Proposed AutoCompress Framework for DNN Structured Pruning}

Given a pretrained DNN or predefined DNN structure, the automatic hyperparameter determination process will decide the per-layer weight pruning rate, and type (and possible combination) of structured pruning scheme per layer. The objective is the maximum reduction in the number of weights or FLOPs, with minimum accuracy loss. 

\subsection{Automatic Process: Generic Flow} \label{sec:generic}

\begin{figure}
    \centering
    \includegraphics[width=0.4\textwidth]{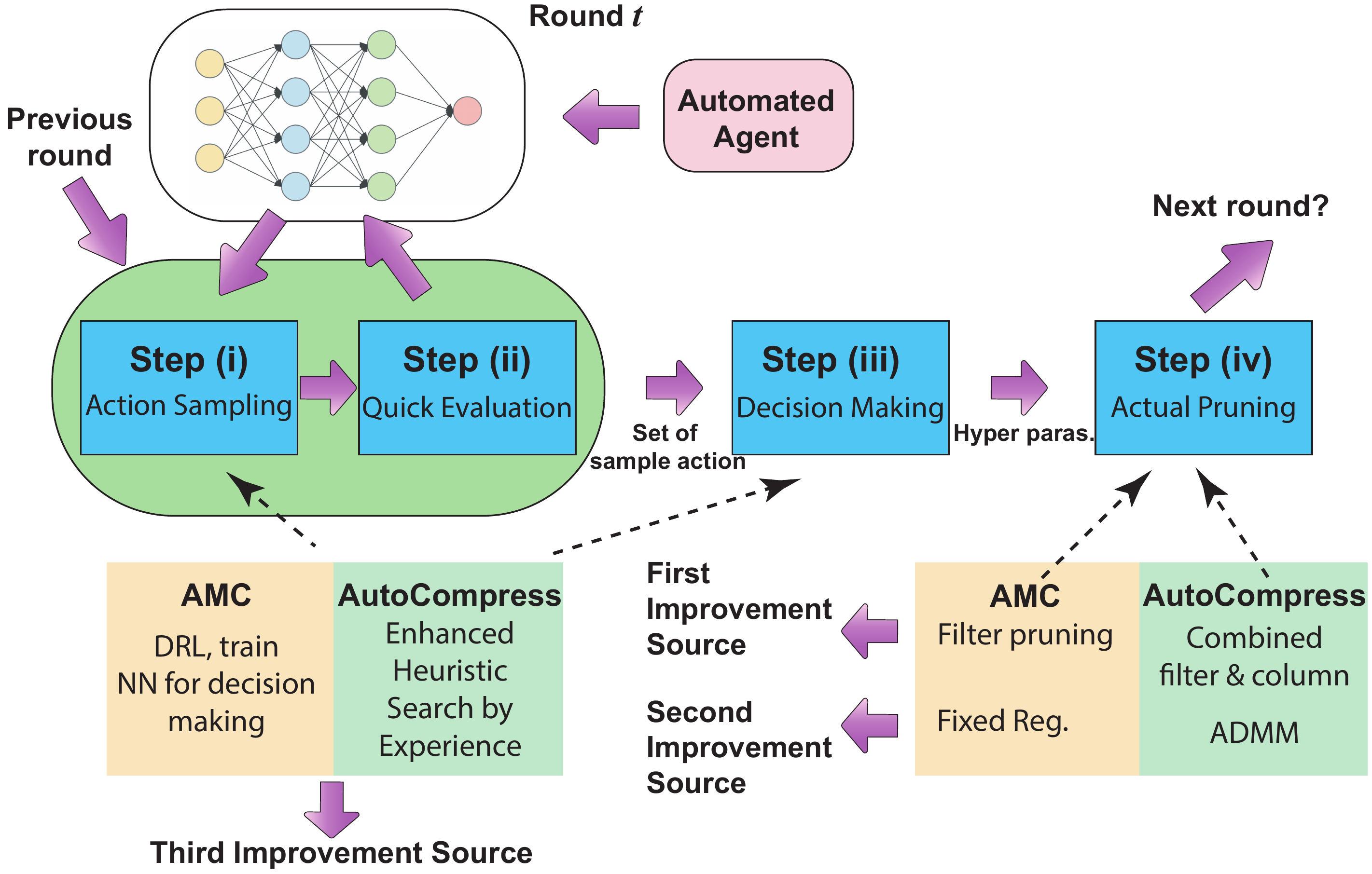}
    \caption{The generic flow of automatic hyperparameter determination framework, and sources of performance improvements.}
    \label{Fig:overview}
    \vspace{-0.1in}
\end{figure}

Figure~\ref{Fig:overview} illustrates the \emph{generic flow} of such automatic process, which applies to both AutoCompress and the prior work AMC. Here we call a sample selection of hyperparamters an ``action" for compatibility with DRL. The flow has the following steps: (i) \emph{action sampling}, (ii) \emph{quick action evaluation}, (iii) \emph{decision making}, and (iv) \emph{actual pruning and result generation}. Due to the high search space of hyperparameters, steps (i) and (ii) should be fast. This is especially important for step (ii), in that we cannot employ the time-consuming, retraining based weight pruning (e.g., fixed regularization \cite{wen2016learning,he2017channel} or ADMM-based techniques) to evaluate the actual accuracy loss. Instead, we can only use simple heuristic, e.g., eliminating a pre-defined portion (based on the chosen hyperparameters) of weights with least magnitudes for each layer, and evaluating the accuracy. This is similar to \cite{he2018amc}. Step (iii) makes decision on the hyperparameter values based on the collection of action samples and evaluations. Step (iv) generates the pruning result, and the optimized (core) algorithm for structured weight pruning will be employed here. Here the algorithm can be more complicated with higher performance (e.g., the ADMM-based one), as it is only performed once in each round.

The overall automatic process is often iterative, and the above steps (i) through (iv) reflect only one round. The reason is that it is difficult to search for high pruning rates in one single round, and the overall weight pruning process will be progressive. This applies to both AMC and  AutoCompress. The number of rounds is 4 - 8 in AutoCompress for fair comparison. Note that AutoCompress supports flexible number of progressive rounds to achieve the maximum weight/FLOPs reduction given accuracy requirement (or with zero accuracy loss).

\subsection{Motivation: Sources of Performance Improvements} \label{sec:motivation}

Based on the generic flow, we identify three sources of performance improvement (in terms of pruning rate, accuracy, etc.) compared with prior work. The \textbf{first} is the \emph{structured pruning scheme}. Our observation is that an effective combination of filter pruning (which is correlated with channel pruning) and column pruning will perform better compared with filter pruning alone (as employed in AMC \cite{he2018amc}). Comparison results are shown in the evaluation section. This is because of the high flexibility in column pruning, while maintaining the hardware-friendly full matrix format in GEMM. The \textbf{second} is the \emph{core algorithm} for structured weight pruning in Step (iv). We adopt the state-of-art ADMM-based weight pruning algorithm in this step. Furthermore, we propose further improvement of a \emph{purification step} on the ADMM-based algorithm taking advantages of the special characteristics after ADMM regularization. In the following two subsections, we will discuss the core algorithm and the proposed purification step, respectively.

The \textbf{third} source of improvement is the underlying principle of action sampling (Step (i)) and decision making (Step (iii)). The DRL-based framework in \cite{he2018amc} adopts an exploration vs. exploitation-based search for action sampling. For Step (iii), it trains a neural network using action samples and fast evaluations, and uses the neural network to make decision on hyperparameter values. Our hypothesis is that DRL is inherently incompatible with the target automatic process, and can be easily outperformed by effective heuristic search methods (such as simulated annealing or genetic algorithm), especially the enhanced versions. More specifically, the DRL-based framework adopted in \cite{he2018amc} is difficult to achieve high pruning rates (the maximum pruning rate in \cite{he2018amc} is only 5$\times$ and is on non-structured pruning), due to the following reasons.

\emph{First}, the sample actions in DRL are generated in a randomized manner, and are evaluated (Step (ii)) using very simple heuristic. As a result, these action samples and evaluation results (rewards) are just rough estimations. When training a neural network and relying on it for making decisions, it will hardly generate satisfactory decisions especially for high pruning rates. \emph{Second}, there is a common limitation of reinforcement learning technique (both basic one and DRL) on optimization problem with constraints \cite{whiteson2011protecting,zhang2016understanding,henderson2018deep}. As pruning rates cannot be set as hard constraints in DRL, it has to adopt a composite reward function with both accuracy loss and weight No./FLOPs reduction. This is the source of issue in controllability, as the relative strength of accuracy loss and weight reduction is very different for small pruning rates (the first couple of rounds) and high pruning rates (the latter rounds). Then there is the paradox of using a single reward function in DRL (hard to satisfy the requirement throughout pruning process) or multiple reward functions (how many? how to adjust the parameters?). \emph{Third}, it is difficult for DRL to support flexible and adaptive number of rounds in the automatic process to achieve the maximum pruning rates. As different DNNs have vastly different degrees of compression, it is challenging to achieve the best weight/FLOPs reduction with a fixed, predefined number of rounds. These can be observed in the evaluation section on the difficulty of DRL to achieve high pruning rates. As these issues can be mitigated by effective heuristic search, we emphasize that an additional benefit of heuristic search is the ability to perform \emph{guided search} based on prior human experience. In fact, the DRL research also tries to learn from heuristic search methods in this aspect for action sampling \cite{osband2016deep,silver2008sample}, but the generality is still not widely evaluated.

\subsection{Core Algorithm for Structured Pruning} \label{sec:core_algorithm}

This work adopts the ADMM-based weight pruning algorithm \cite{zhang2018systematic,zhang2018adam} as the core algorithm, which generates state-of-art results in both non-structured and structured weight pruning. Details are in \cite{zhang2018systematic,zhang2018adam,boyd2011distributed,suzuki2013dual}. The major step in the algorithm is \emph{ADMM regularization}.
Consider a general DNN with loss function $f(\{{\bf{W}}_{i}\},\{{\bf{b}}_{i}\})$, where ${\bf{W}}_{i}$ and ${\bf{b}}_{i}$ correspond to the collections of weights and biases in layer $i$, respectively. The overall (structured) weight pruning problem is defined as
\begin{equation}
\underset{\{{\bf{W}}_{i}\},\{{\bf{b}}_{i}\}}{\text{minimize}}
\ f(\{{\bf{W}}_{i}\},\{{\bf{b}}_{i}\}),\ \  \text{subject to}\  {\bf{W}}_{i} \in \mathcal{S}_{i},\ \text{for all}\ i;
\end{equation}
where $\mathcal{S}_{i}$ reflects the requirement that remaining weights in layer $i$ satisfy predefined ``structures". Please refer to \cite{wen2016learning,he2017channel} for more details.

By defining (i) indicator functions 
$g_{i}({\bf{W}}_{i})=
\begin{cases}
 0 & \text { if } {\bf{W}}_{i}\in {\mathcal{S}}_{i} \\ 
 +\infty & \text { otherwise}
\end{cases}$, 
(ii) incorporating auxiliary variable ${\bf{Z}}_{i}$ and dual variable ${\bf{U}}_{i}$, (iii) adopting augmented Lagrangian \cite{boyd2011distributed}, the ADMM regularization decomposes the overall problem into two subproblems, and iteratively solved them until convergence. The first subproblem is $\displaystyle \underset{ \{{\bf{W}}_{i}\},\{{\bf{b}}_{i} \}}{\text{minimize}}
\ f \big( \{{\bf{W}}_{i} \}_{i=1}^N, \{{\bf{b}}_{i} \}_{i=1}^N \big)+\sum_{i=1}^{N} \frac{\rho_{i}}{2}  \| {\bf{W}}_{i}-{\bf{Z}}_{i}^{k}+{\bf{U}}_{i}^{k} \|_{F}^{2}.$ It can be solved using current gradient descent techniques and solution tools for DNN training. The second subproblem is $\displaystyle \underset{ \{{\bf{Z}}_{i} \}}{\text{minimize}}
\ \ \ \sum_{i=1}^{N} g_{i}({\bf{Z}}_{i})+\sum_{i=1}^{N} \frac{\rho_{i}}{2} \| {\bf{W}}_{i}^{k+1}-{\bf{Z}}_{i}+{\bf{U}}_{i}^{k} \|_{F}^{2}$, which can be optimally solved as Euclidean mapping. 

Overall speaking, ADMM regularization is a dynamic regularization where the regularization target is dynamically adjusted in each iteration, without penalty on all the weights. This is the reason that ADMM regularization outperforms prior work of fixed $L_1$, $L_2$ regularization or projected gradient descent (PGD). To further enhance the convergence rate, the \emph{multi-$\rho$ method} \cite{ye2018progressive} is adopted in ADMM regularization, where the $\rho_i$ values will gradually increase with ADMM iterations.

\vspace{-0.05in}
\subsection{Purification and Unused Weights Removal} \label{sec:purification}

After ADMM-based structured weight pruning, we propose the purification and unused weights removal step for further weight reduction without accuracy loss. First, as also noticed by prior work \cite{he2017channel}, a specific filter in layer $i$ is responsible for generating one channel in layer $i+1$. As a result, removing the filter in layer $i$ (in fact removing the batch norm results) also results in the removal of the corresponding channel, thereby achieving further weight reduction. Besides this straightforward procedure, there is further margin of weight reduction based on the characteristics of ADMM regularization. As ADMM regularization is essentially a dynamic, $L_2$-norm based regularization procedure, there are a large number of non-zero, small weight values after regularization. Due to the non-convex property in ADMM regularization, \textbf{our observation} is that removing these weights can maintain the accuracy or even slightly improve the accuracy occasionally. As a result, we define two thresholds, a \emph{column-wise threshold} and a \emph{filter-wise threshold}, for each DNN layer. When the $L_2$ norm of a column (or filter) of weights is below the threshold, the column (or filter) will be removed. Also the corresponding channel in layer $i+1$ can be removed upon filter removal in layer $i$. Structures in each DNN layer will be maintained after this purification step.

These two threshold values are layer-specific, depending on the relative weight values of each layer, and the sensitivity on overall accuracy. They are hyperparameters to be determined for each layer in the AutoCompress framework, for maximum weight/FLOPs reduction without accuracy loss.

\subsection{The Overall AutoCompress Framework for Structured Weight Pruning and Purification}

\begin{figure}
    \setlength{\belowcaptionskip}{-0.7cm}
    \centering
    \includegraphics[width=0.4\textwidth]{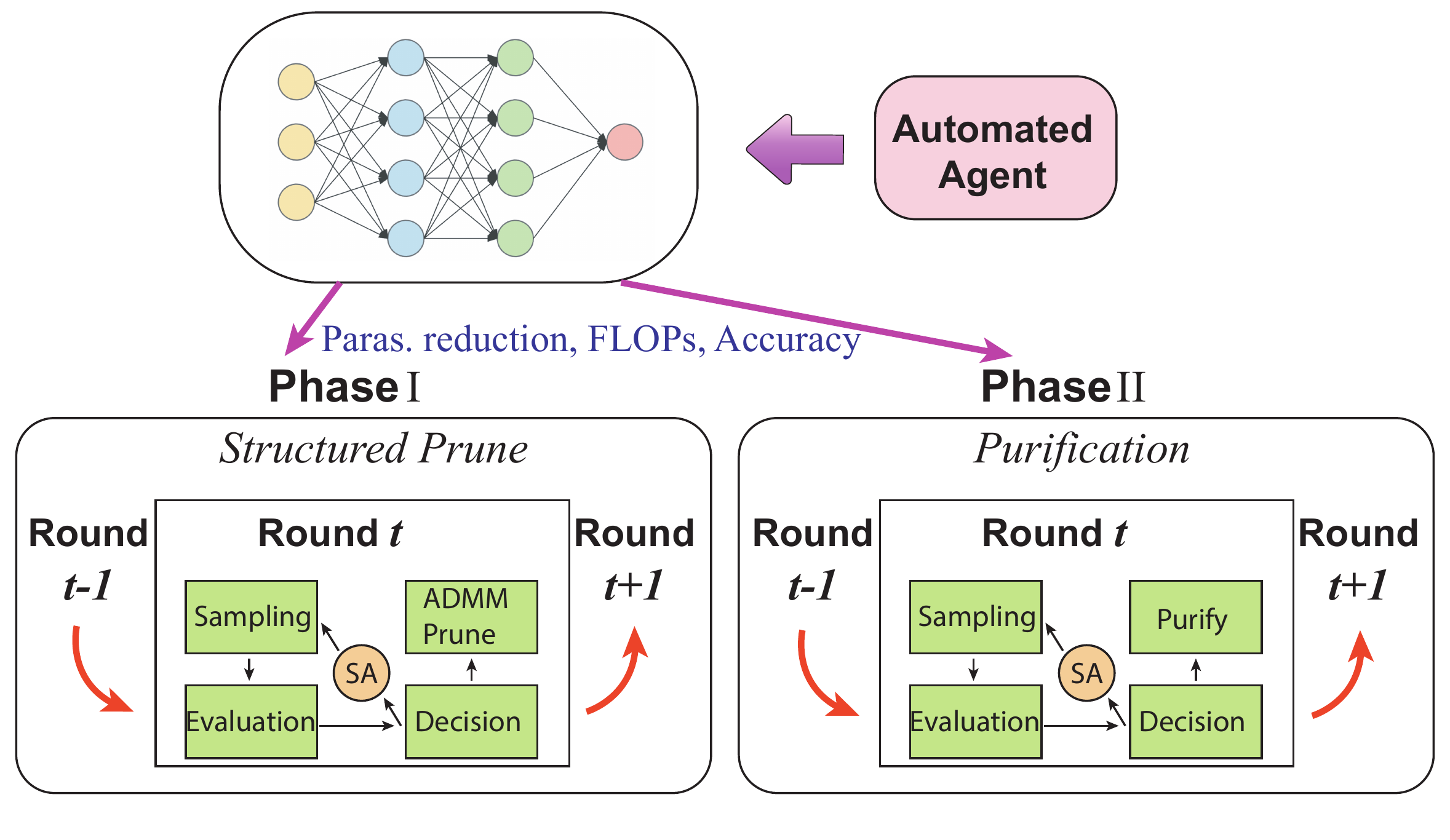}
    \caption{Illustration of the AutoCompress framework.}
    \label{Fig:framework}
\end{figure}

In this section, we discuss the AutoCompress framework based on the enhanced, guided heuristic search method, in which the automatic process determines per-layer weight pruning rates, structured pruning schemes (and combinations), as well as hyperparameters in the purification step (discussed in Section~\ref{sec:purification}). The overall framework has two phases as shown in Figure \ref{Fig:framework}: \emph{Phase I} for structured weight pruning based on ADMM, and \emph{Phase II} for the purification step. Each phase has multiple progressive rounds as discussed in Section~\ref{sec:generic}, in which the weight pruning result from the previous round serves as the starting point of the subsequent round. We use Phase I as illustrative example, and Phase II uses the similar steps.

The AutoCompress framework supports flexible number of progressive rounds, as well as hard constraints on the weight or FLOPs reduction. In this way, it aims to achieve the maximum weight or FLOPs reduction while maintaining accuracy (or satisfying accuracy requirement). For each round $t$, we set the overall reduction in weight number/FLOPs to be a factor of 2 (with a small variance), based on the result from the previous round. In this way, we can achieve around $4\times$ weight/FLOPs reduction within 2 rounds, already outperforming the reported structured pruning results in prior work \cite{he2018amc}.

We leverage a classical heuristic search technique \emph{simulated annealing} (SA), with enhancement on \emph{guided search} based on prior experience. The enhanced SA technique is based on the observation that a DNN layer with more number of weights often has a higher degree of model compression with less impact on overall accuracy. The basic idea of SA is in the search for actions: When a perturbation on the candidate action results in better evaluation result (Step (ii) in Figure~\ref{Fig:overview}), the perturbation will be accepted; otherwise the perturbation will be accepted with a probability depending on the degradation in evaluation result, as well as a temperature $T$. The reason is to avoid being trapped in local minimum in the search process. The temperature $T$ will gradually decrease during the search process, in analogy to the physical ``annealing" process.
        \begin{algorithm}\footnotesize
        \caption{\footnotesize AutoCompress Framework for Structured Weight Pruning (similar process also applies to purification).}
        \hspace*{0.02in} {\bf REQUIRE}: Initial (unpruned) DNN model or DNN structure.
        \begin{algorithmic}
        \For{each progressive round $t$}
        \State Initialize the action $A_t^0$ with partitioning of structured prun-
        \State ing schemes and pruning rate $\approx C_t$, satisfying the heuristic 
        \State constraint.
            \While{$T$ $>$ stop temperature}
                \For{iteration $i$}
                    \State Generate \emph{perturbation} (magnitude decreases with $T$)
                    \State on action, satisfying the heuristic constraint.
                    \State Perform fast evaluation on the perturbation.
                    \If{better evaluation result (higher accuracy)}
                        \State Accept the \emph{perturbation}.
                    \Else
                        \State Accept with probability $e^{-\frac{\Delta E}{T}}$, where $\Delta E$ is 
                        \State increase in evaluation cost (accuracy loss).
                    \EndIf
                \EndFor
            \State Cool down $T\leftarrow \eta \cdot T$.
            \EndWhile
                    \State The action outcome becomes the decision of hyperparameter 
                    \State values.
                    \State Perform ADMM-based structured pruning to generate prun-
                    \State ing result, for the next round.
        \EndFor
        \end{algorithmic}
        \end{algorithm}
        
Given the overall pruning rate $C_t\approx 2$ (on weight No. or FLOPs) in the current round, we initialize a randomized action $A_t^0$ using the following process: i) order all layers based on the number of remaining weights, ii) assign a randomized pruning rate (and partition between filter and column pruning schemes) for each layer, satisfying that a layer with more weights will have no less pruning rate, and iii) normalize the pruning rates by $C_t$. We also have a high initialized temperature $T$. We define \emph{perturbation} as the change of weight pruning rates (and portion of structured pruning schemes) in a subset of DNN layers. The perturbation will also satisfy the requirement that the layer will more remaining weights will have a higher pruning rate. The result evaluation is the fast evaluation introduced in Section~\ref{sec:generic}. The acceptance/denial of action perturbation, the degradation in temperature $T$, and the associated reduction in the degree of perturbation with $T$ follow the SA rules until convergence. The action outcome will become the decision of hyperparameter values (Step (iii), this is different from DRL which trains a neural network). The ADMM-based structured pruning will be adopted to generate pruning result (Step (iv)), possibly for the next round until final result.

\section{Evaluation, Experimental Results, and Discussions} \label{sec:evaluation}

\begin{table*}\tiny
    \centering
    \caption{Comparison on pruning approaches using VGG-16 on CIFAR-10 Dataset}
    \renewcommand{\arraystretch}{1}
    \setlength{\tabcolsep}{7mm}{
        \resizebox{18cm}{!}{
            \begin{tabular}{c c c c c c} 
    					\hline\hline
                    	\multirow{2}{*}{} & \multirow{2}{*}{\makecell{Method}} & \multirow{2}{*}{\makecell{Accuracy}} & \multirow{2}{*}{\makecell{CONV \\ Params Rt.}} & \multirow{2}{*}{\makecell{CONV \\ FLOPs Rt.}} & \multirow{2}{*}{\makecell{Inference time}}  \\  \\
    					\hline
    					\bf{Original VGG-16} &   & 93.7\% & 1.0$\times$ & 1.0$\times$ & 14ms \\
    					\hline
    					\multirow{3}{*}{\makecell{Filter Pruning}} 
    					&2PFPCE \cite{min20182pfpce} & 92.8\% & $4\times$ & N/A & N/A\\
    					&2PFPCE~\cite{min20182pfpce} & 91.0\% & 8.3$\times$ & N/A & N/A\\
    					&ADMM, manual hyper. determ. & 93.48\% & 9.3$\times$ & 2.1$\times$ & 7.1ms\\
    					\hline
    					\multirow{3}{*}{\makecell{Auto Filter Pruning}} 
    					&ADMM-based, enhanced SA & 93.22\% & 13.7$\times$ & 3.1$\times$ & 4.8ms\\
    					& Train-From-Scratch & 93.19\% & 13.7$\times$ & 3.1$\times$ & 4.8ms \\
    					&ADMM-based, enhanced SA & 88.78\% & 47.4$\times$ & 14.0$\times$ & 1.7ms\\
    					\hline
    					\multirow{3}{*}{\makecell{Combined Structured Pruning}} 
    					& ADMM, manual hyper. determ. & 93.26\% & 44.3$\times$ & 8.1$\times$ & 2.9ms \\
    					& Full \textbf{AutoCompress} & \textbf{93.21\%} & 52.2$\times$ & 8.8$\times$ & 2.7ms\\
    					& Train-From-Scratch & 91.4\% & 52.2$\times$ & 8.8$\times$ & 2.7ms\\

    					\hline
            \end{tabular}
        }
    }
    \vspace{-1mm}
    \label{table:results_VGGNet}
\end{table*}

\begin{table*}[htb]\tiny
    \centering
    \caption{Comparison on pruning approaches using ResNet-18 (ResNet-50 in NISP and AMC) on CIFAR-10 Dataset}
    \renewcommand{\arraystretch}{1}
\setlength{\tabcolsep}{7mm}{
    \resizebox{18cm}{!}{
        \begin{tabular}{cccccc} 
					\hline\hline
                    \multirow{2}{*}{} & \multirow{2}{*}{Method} & \multirow{2}{*}{\makecell{Accuracy}} & \multirow{2}{*}{\makecell{CONV \\ Params Rt.}} & \multirow{2}{*}{\makecell{CONV \\ FLOPs Rt.}} & \multirow{2}{*}{\makecell{Inference time}}  \\  \\
					\hline
					\bf{Original ResNet-18} &   & 93.9\% & 1.0$\times$ & 1.0$\times$ & 11ms \\
					\hline
					\multirow{2}{*}{\makecell{Filter Pruning}} 
					&NISP \cite{yu2018nisp} & 93.2\% & $1.7\times$ & N/A & N/A\\
					&ADMM, manual hyper. determ. & 93.9\% & 5.2$\times$ & 2.7$\times$ & 4.2ms\\
					\hline
					\multirow{3}{*}{\makecell{Auto Filter Pruning}} 
					&AMC \cite{he2018amc} & 93.5\% & 1.7$\times$ & N/A & N/A\\
					&ADMM-based, enhanced SA & 93.91\% & 8.0$\times$ & 4.7$\times$ & 2.4ms\\
					& Train-From-Scratch & 93.89\% & 8.0$\times$ & 4.7$\times$ & 2.4ms \\
					\hline
					\multirow{5}{*}{\makecell{Combined Structured Pruning}} 
					& ADMM, DRL hyper. determ. & 93.55\% & 11.8$\times$ & 3.8$\times$ & 4.7ms \\
					& ADMM, manual hyper. determ. & 93.69\% & 43.3$\times$ & 9.6$\times$ & 1.9ms \\
					& Full \textbf{AutoCompress} & 93.43\% & 61.2$\times$ & 13.3$\times$ & 1.3ms \\
					& Full \textbf{AutoCompress} & \textbf{93.81\%} & 54.2$\times$ & 12.2$\times$ & 1.45ms\\
					& Train-From-Scratch & 91.88\% & 54.2$\times$ & 12.2$\times$ & 1.45ms \\
					\hline
        \end{tabular}
    }
    }
    \vspace{-1mm}
    \label{table:results_ResNet}
\vspace{-0.15in}
\end{table*}

\textbf{Setup:} The effectiveness of AutoCompress is evaluated on VGG-16 and ResNet-18 on CIFAR-10 dataset, and VGG-16 and ResNet-18/50 on ImageNet dataset. We focus on the structured pruning on CONV layers, which are the most computationally intensive layers in DNNs and the major storage in state-of-art DNNs such as ResNet. 
In this section we focus on the objective function of reduction in the number of weight parameters, and leave the objective in the amount of computation (FLOPs) in Supplementary Materials. 
The implementations are based on PyTorch \cite{paszke2017automatic}.
For structured pruning, we support (i) filter pruning only, and (ii) combined filter and column pruning, both supported in ADMM-based algorithm and AutoCompress framework.
In the ADMM-based structured pruning algorithm, the number of epochs in each progressive round is 200, which is lower than the prior iterative pruning heuristic \cite{han2015learning}. 
We use an initial penalty parameter $\vec{\rho} = 10^{-4}$ for ADMM and initial learning rate $10^{-3}$. The ADAM \cite{kingma2014adam} optimizer is utilized. 
In the SA setup, we use cooling factor $\eta = 0.7$ and Boltzmann’s constant $k = 10^{-3}$. The initial probability of accepting high energy (bad) moves is set to be relatively high.

\textbf{Models and Baselines:}
We aim at fair and comprehensive evaluation on the effectiveness of three sources of performance improvements discussed in Section \ref{sec:motivation}.
Besides the original, unpruned DNN models, we compare with a set of prior baseline methods. Perhaps for software implementation convenience, almost all baseline methods we can find focus on filter/channel pruning. For fair comparison, we also provide pruning results on ADMM-based filter pruning with manual hyperparameter determination. This case is only different from prior work by a single source of performance improvement -- the core algorithm using ADMM. We also show the results on ADMM-based filter pruning with enhanced SA-based hyperparameter determination, in order to show the effect of an additional source of improvement.

Beyond filter pruning only, we show the combined structured pruning results using ADMM to demonstrate the last source of performance improvement. We provide results on manual, our crafted DRL-based, and enhanced SA-based hyperparameter determination for fair comparison, the last representing the full version of AutoCompress. We provide the inference time of the pruned models using the latest Qualcomm Adreno 640 GPU in Samsung Galaxy S10 smartphone.
The results clearly demonstrate the actual acceleration using the combined structured pruning. Note that our mobile DNN acceleration framework is a compiler assisted, strong framework by itself. For the original VGG-16 and ResNet-18 (without pruning) on CIFAR-10, it achieves 14ms and 11ms end-to-end inference times, respectively, on the Adreno 640 mobile GPU.
For the original VGG-16 and ResNet-50 on ImageNet, it achieves 95ms and 48ms inference times, respectively. All these results, as starting points, outperform current DNN acceleration frameworks like TensorFlow-Lite \cite{TensorFlow-Lite} and TVM \cite{chen2018tvm}.

Recent work \cite{liu2018rethinking} points out an interesting aspect. When one trains from scratch based on the structure (not using weight values) of a pruned model, one can often retrieve the same accuracy as the model after pruning. We incorporate this ``Train-From-Scratch" process based on the results of filter pruning and combined filter and column pruning (both the best results using the enhanced SA-based search). We will observe whether accuracy can be retrieved.

Through extensive experiments, we conclude that AutoCompress is the key to achieve ultra-high pruning rates on the number of weights and FLOPs that cannot be achieved before, while DRL cannot compete with human experts to achieve high structured pruning rates.

\subsection{Results and Discussions on CIFAR-10 Dataset}

Table \ref{table:results_VGGNet} illustrates the comparison results on VGG-16 for CIFAR-10 dataset, while Table \ref{table:results_ResNet} shows the results on ResNet-18 (ResNet-50 for some baselines but their accuracy is similar to our ResNet-18). The objective function of our AutoCompress framework is reducing the number of weight parameters (please refer to Supplementary Materials for the cases when FLOPs reduction is set as objective), but the FLOPs reduction results are also reported. 

From the two tables we have the following conclusions. \textbf{First}, for filter/channel pruning only using manual hyperparameter determination, our method outperforms prior work 2PFPCE, NISP and AMC (both in accuracy and in pruning rate). As no other sources of improvement are exploited, this improvement is attributed to the ADMM-based algorithm equipped with purification. \textbf{Second}, the combined structured pruning outperforms filter-only pruning in both weight reduction and FLOPs reduction. For manual hyperparameter determination, the combined structured pruning enhances from 9.3$\times$ pruning rate to 44.3$\times$ in VGG-16, and enhances from 5.2$\times$ to 43.3$\times$ in ResNet-18. If we aim at the same high pruning rate for filter-only pruning, it suffers a notable accuracy drop (e.g., 88.78\% accuracy at 47.4$\times$ pruning rate for VGG-16). \textbf{Third}, the enhanced SA-based hyperparameter determination outperforms DRL and manual counterparts. As can be observed in the two tables, the full AutoCompress achieves a moderate improvement in pruning rate compared with manual hyperparameter optimization, but significantly outperforms DRL-based framework (all other sources of improvement are the same). This demonstrates the statement that DRL is not compatible with ultra-high pruning rates. For relatively small pruning rates, it appears that DRL can hardly outperform manual process as well, as the improvement over 2PFPCE is less compared with the improvement over AMC.

With all sources of performance improvements effectively exploited, the full AutoCompress framework achieves 15.3$\times$ improvement in weight reduction compared with 2PFPCE and 33$\times$ improvement compared with NISP and AMC, under the same (or higher for AutoCompress) accuracy. When accounting for the different number of parameters in ResNet-18 and ResNet-50 (NISP and AMC), the improvement can be even perceived as 120$\times$. It demonstrates the significant performance of our proposed AutoCompress framework, and also implies that the high redundancy of DNNs on CIFAR-10 dataset has not been exploited in prior work. Also the measured inference speedup on mobile GPU validates the effectiveness of the combined pruning scheme and our proposed AutoCompress framework.

Moreover, there are some interesting results on ``Train-From-Scratch" cases, in response to the observations in \cite{liu2018rethinking}. When ``Train-From-Scratch" is performed based the result of filter-only pruning, it can recover the similar accuracy. The insight is that filter/channel pruning is similar to finding a smaller DNN model. In this case, the main merit of AutoCompress framework is to discover such DNN model, especially corresponding compression rates in each layer, and our method still outperforms prior work. On the other hand, when ``Train-From-Scratch" is performed based on the result of combined structured pruning, \underline{the accuracy CANNOT be recovered}. This is an interesting observation. The underlying insight is that the combined pruning is not just training a smaller DNN model, but with adjustments of filter/kernel shapes. In this case, the pruned model represents a solution that cannot be achieved through DNN training only, even with detailed structures already given. In this case, weight pruning (and the AutoCompress framework) will be more valuable due to the importance of training from a full-sized DNN model.

\subsection{Results and Discussions on ImageNet Dataset}

In this subsection, we show the application of AutoCompress on ImageNet dataset, and more comparison results with filter-only pruning (equipped by ADMM-based core algorithm and SA-based hyperparameter determination). This will show the first source of improvement. 
Table \ref{vggimagenet} and Table \ref{resnetimagenet} show the comparison results on VGG-16 and ResNet-18 (ResNet-50) structured pruning on ImageNet dataset, respectively.
We can clearly see the advantage of AutoCompress over prior work, such as \cite{he2017channel} (filter pruning with manual determination), AMC \cite{he2018amc} (filter pruning with DRL), and ThiNet \cite{luo2017thinet} (filter pruning with manual determination).
We can also see the advantage of AutoCompress over manual hyperparameter determination (both combined structured pruning with ADMM-based core algorithm), improving from 2.7$\times$ to 3.3$\times$ structured pruning rates on ResNet-18 (ResNet-50) under the same (Top-5) accuracy.
Finally, the full AutoCompress also outperforms filter pruning only (both ADMM-based core algorithm and SA-based hyperparameter determination), improvement from 3.8$\times$ to 6.4$\times$ structured pruning rates on VGG-16 under the same (Top-5) accuracy. This demonstrates the advantage of combined filter and column pruning compared with filter pruning only, when the other sources of improvement are the same. Besides, our filter-only pruning results also outperform prior work, demonstrating the strength of proposed framework.

\begin{table}[htb]\scriptsize
    \renewcommand\arraystretch{0.8}
    \centering
    \begin{tabular}{llll}
    \toprule
    \cmidrule(r){1-3}
    Method & Top-5 Acc. Loss & Params Rt. & Objective \\ 
    \text{Filter \cite{he2017channel}} & 1.7\% & $\approx$ 4$\times$ & N/A \\
    AMC \cite{he2018amc} & 1.4\% & $\approx$ 4$\times$ & N\/A \\
    Filter pruning, ADMM, SA & 0.6\% & 3.8$\times$ & Params\# \\
    Full \textbf{AutoCompress} & 0.6\% & \textbf{6.4$\times$} & Params\# \\
    \bottomrule
  \end{tabular}
  \vspace{0.05in}
    \caption{Comparison results on VGG-16 for the ImageNet dataset.}
    \label{vggimagenet}
    
\end{table}
\vspace{-0.25in}

\begin{table}[htb]\scriptsize
    \setlength{\belowcaptionskip}{-0.5cm} 
    \renewcommand\arraystretch{0.8}
    \centering
    \setlength{\tabcolsep}{1.6mm}{
    \begin{tabular}{c c c c}
    \toprule
    \cmidrule(r){1-3}
    Method & Top-5 Acc. Loss & Params Rt. & Objective \\ 
    ThiNet-50 \cite{luo2017thinet} & 1.1\% & $\approx$ 2$\times$ & N/A \\
    ThiNet-30 \cite{luo2017thinet} & 3.5\% & $\approx$ 3.3$\times$ & N/A \\
        Filter pruning, ADMM, SA & 0.8\% & 2.7$\times$ & Params\# \\
    Combined pruning, ADMM, manual & 0.1\% & 2.7$\times$ & N/A \\
    Full \textbf{AutoCompress} & 0.1\% & \textbf{3.3$\times$} & Params\# \\
    \bottomrule
  \end{tabular}
  }
    \caption{Comparison results on ResNet-18 (ResNet-50) for the ImageNet dataset.}
    \label{resnetimagenet}
\end{table}

\begin{table}[htb]\scriptsize
    \renewcommand\arraystretch{0.8}
    \centering
    \setlength{\tabcolsep}{3mm}{
    \begin{tabular}{c c c c c}
    \toprule
    \cmidrule(r){1-3}
    Method & Top-5 Acc. Loss & Params Rt. & Objective \\ 
    AMC \cite{he2018amc} & 0\% & 4.8$\times$ & N/A \\
    ADMM, manual hyper. & 0\% & 8.0$\times$ & N/A \\
    Full \textbf{AutoCompress} & 0\% & \textbf{9.2$\times$} & Params\# \\
    Full \textbf{AutoCompress} & 0.7\% & \textbf{17.4$\times$} & Params\# \\
    \bottomrule
  \end{tabular}
  }
  \vspace{0.05in}
    \caption{Comparison results on non-structured weight pruning on ResNet-50 using ImageNet dataset.}
    \label{wrap-tab:1}
    \vspace{-0.1in}
\end{table}

Last but not least, the proposed AutoCompress framework can also be applied to \underline{non-structured pruning}. For non-structured pruning on ResNet-50 model for ImageNet dataset, AutoCompress results in 9.2$\times$ non-structured pruning rate on CONV layers without accuracy loss (92.7\% Top-5 accuracy), which outperforms manual hyperparameter optimization with ADMM-based pruning (8$\times$ pruning rate) and prior work AMC (4.8$\times$ pruning rate).

\section{Conclusion}

This work proposes AutoCompress, an automatic structured pruning framework with the following key performance improvements: (i) effectively incorporate the combination of structured pruning schemes in the automatic process; (ii) adopt the state-of-art ADMM-based structured weight pruning as the core algorithm, and propose an innovative additional purification step for further weight reduction without accuracy loss; and (iii) develop effective heuristic search method enhanced by experience-based guided search, replacing the prior deep reinforcement learning technique which has underlying incompatibility with the target pruning problem.
Extensive experiments on CIFAR-10 and ImageNet datasets demonstrate that AutoCompress is the key to achieve ultra-high pruning rates on the number of weights and FLOPs that cannot be achieved before.
\small{\bibliography{reference}}
\bibliographystyle{aaai}


\newpage
\onecolumn
\section*{Supplementary Materials}

\subsection*{Comparison between two objective functions (weight parameter reduction and FLOPs reduction)}

\begin{table}[htb]\scriptsize
    \centering
    \renewcommand{\arraystretch}{0.8}
    \setlength{\tabcolsep}{7mm}{
    \resizebox{18cm}{!}{
    
    \begin{tabular}{l l l l l}
    \toprule
    \cmidrule(r){1-3}
    Method     & Acc.   & Params Rt.  & FLOPs Rt.  & Objective \\ 
    \midrule
    2PFPCE \cite{min20182pfpce} & 92.8\% & 4$\times$ & N/A & N/A \\
    \midrule
    2PFPCE~\cite{min20182pfpce} & 91.0\% & 8.3$\times$ & N/A & N/A \\
    \midrule
    Combined struct. prun., ADMM, manual determ. & 93.26\% & 44.3$\times$ & 8.1$\times$ & N/A \\
    \midrule
    Full \textbf{AutoCompress} & \textbf{93.21\%} & 52.2$\times$ & 8.8$\times$ & Params\# \\
    \midrule
    Full \textbf{AutoCompress} & 92.72\% & \textbf{61.1$\times$} & 10.6$\times$ & Params\# \\
    \midrule
    Full \textbf{AutoCompress} & 92.65\% & 59.1$\times$ & \textbf{10.8$\times$} & FLOPs\# \\
    \midrule
    Full \textbf{AutoCompress} & 92.79\% & 51.3$\times$ & 9.1$\times$ & FLOPs\# \\
    \bottomrule
  \end{tabular}
  }
  }
    \vspace{0.1in}
    \caption{Comparison results on VGG-16 for CIFAR-10 dataset.}
    \label{vggcifar10}
\end{table}

\vspace{-0.2in}
\begin{table}[htb]\scriptsize
    \centering
    \renewcommand{\arraystretch}{0.8}
    \setlength{\tabcolsep}{7mm}{
    \resizebox{18cm}{!}{
    \begin{tabular}{p{4cm}llll}
    \toprule
    \cmidrule(r){1-3}
    Method     & Acc.     & Params Rt & FLOPs Red. &  Objective \\
    \midrule
    AMC \cite{he2018amc} & 93.5\% & 1.7$\times$ & N/A & N/A \\
    \midrule
    NISP \cite{yu2018nisp} & 93.2\% & $1.7\times$ & N/A & N/A\\
    \tiny{Combined struct. prun., ADMM, DRL determ.} & 93.55\% & 11.8$\times$ & 3.8$\times$ & Params\# \\
    \midrule
    \tiny{Combined struct. prun., ADMM, manual determ.} & 93.69\% & 43.3$\times$ & 9.6$\times$ & N/A \\
    \midrule
    Full \textbf{AutoCompress} & 93.75\% & 55.6$\times$ & 12.3$\times$ & Params\# \\
    \midrule
    Full \textbf{AutoCompress} & 93.43\% & 61.2$\times$ & 13.3$\times$ & Params\# \\
    \midrule
    Full \textbf{AutoCompress} & 92.98\% & \textbf{80.8$\times$} & \textbf{17.2$\times$} & Params\# \\
    \midrule
    Full \textbf{AutoCompress} & \textbf{93.81\%} & 54.2$\times$ & 12.2$\times$ & FLOPs\# \\
    \bottomrule
  \end{tabular}
  }
  }
  \vspace{0.1in}
    \caption{Comparison results on ResNet-18 (ResNet-50 in AMC) for CIFAR-10 dataset.}
    \label{resnetcifar10}
    \vspace{-0.2in}
\end{table}

\begin{figure}[h!]
    \centering
    \includegraphics[width=0.5\textwidth]{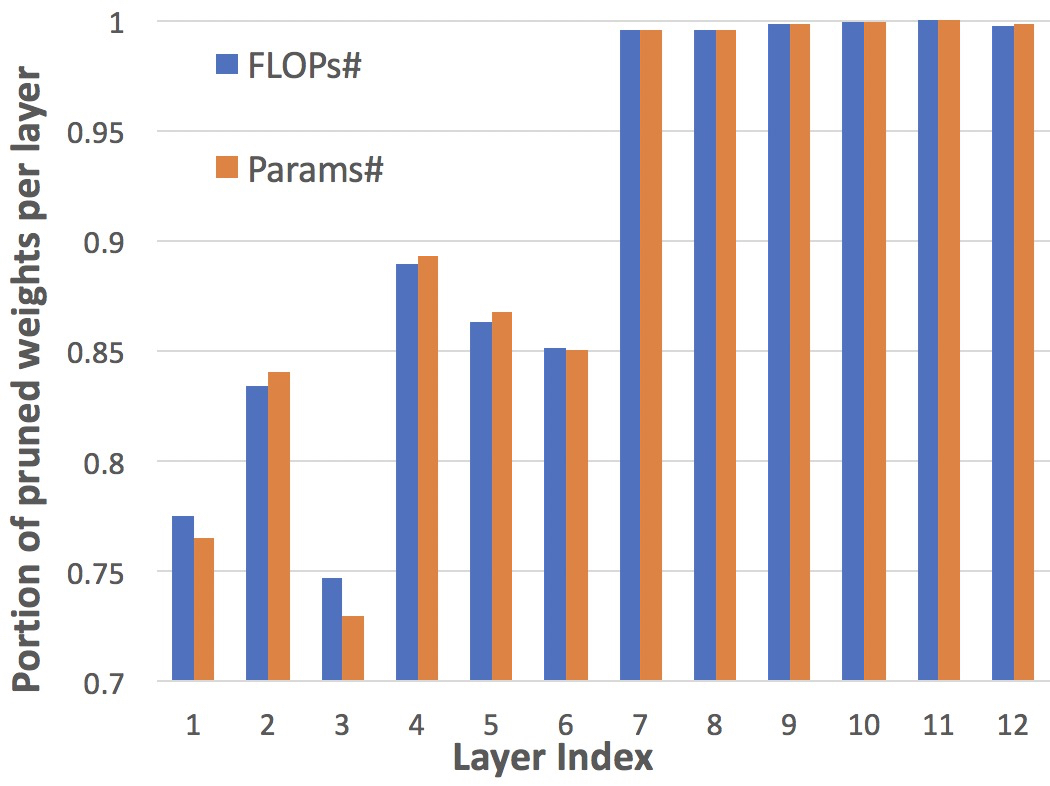}
    \caption{Detailed per-layer portion of pruned weights on VGG-16 for CIFAR-10 under two objective functions Params\# and FLOPs\#.}
    \label{Fig:vgg_row_v3}
\end{figure}

We compare between two objectives: weight No. and FLOPs reductions, in the AutoCompress framework.
Detailed results on VGG-16 and ResNet-18 models for CIFAR-10 dataset are shown in Table 6 and Table 7, respectively.
Moreover, Figure \ref{Fig:vgg_row_v3} illustrates the portion of pruned weights per layer on VGG-16 for CIFAR-10 by Params\# and FLOPs\# search objectives. One can only observe slight difference in the portion of pruned weights per layer (also the tables show that parameter count reduction and FLOPs reduction are highly correlated). This is because further weight reduction in the first several layers of VGG-16 and ResNet-18 models (and many other DNN models alike) results in significant accuracy degradation. The somewhat convergence in results using two objectives seems to be one interesting feature under ultra-high pruning rates. It also provides some hints that the observations (and conclusions) under relatively low pruning rates may not still hold under ultra-high pruning rates.

\end{document}